\documentclass[11pt]{article}

\usepackage[preprint]{acl}

\usepackage{times}
\usepackage{latexsym}
\usepackage[table]{xcolor} 
\usepackage[T1]{fontenc}

\usepackage[utf8]{inputenc}

\usepackage{microtype}

\usepackage{inconsolata}

\usepackage{graphicx}

\title{Modelling the Morphology of Verbal Paradigms: A Case Study in the Tokenization of Turkish and Hebrew}

\author{Giuseppe Samo \\
  Idiap Research Institute \\
  \texttt{giuseppe.samo@idiap.ch} \\\And
  Paola Merlo \\
  Idiap Research Institute  \\
  University of Geneva \\
  \texttt{paola.merlo@idiap.ch} \\}

\begin{document}
\maketitle
\begin{abstract}
We investigate how transformer models represent complex verb paradigms in Turkish and Modern Hebrew, concentrating on how tokenization strategies shape this ability. Using the Blackbird Language Matrices task on natural data, we show that for Turkish --with its transparent morphological markers-- both monolingual and multilingual models succeed, either when tokenization is atomic or when it breaks words into small subword units. For Hebrew, instead, monolingual and multilingual models diverge. A multilingual model using character-level tokenization fails to capture the language non-concatenative morphology, but a monolingual model with morpheme-aware segmentation performs well. Performance improves on more synthetic datasets, in all models. 
\end{abstract}

\section{Introduction}

\label{introduction}
While language models excel at capturing distributional information at the token and sentence level \citep{warstadt-etal-2019-investigating,linzenbaroni21,gautam2024subject}, their ability to generalize over paradigmatic phenomena, such as verb alternations \citep{Levin93,kastner2019templatic} and systematic patterns of verbal inflection remains less well understood \citep{yi-etal-2022-probing,proietti-etal-2022-bert,samo-etal-2023-blm}. The tokenisation step is an important aspect in understanding how a model treats verb alternations and paradigms in general, as tokenization shapes the internal representations in language models. 

Paradigms capture the relational and systematic nature of linguistic elements, making  variation meaningful and predictable within a broader structural framework \citep{Setzke+2024+73+94,bobaljik2015suppletion}.
Consider a simple case: in many languages, causative verbs (verbs whose meaning implies that an actor caused the event described by the main verb) typically exhibit two voices, a transitive (\textsc{T}) and an intransitive (\textsc{I}) alternant. 

Languages differ in the way they encode the voices of a paradigm \citep{haspelmath2014coding, samardvzic2018probability}. In English, neither alternant is morphologically marked (e.g., \textit{the chef melts\textsubscript{T} the butter} vs. \textit{the butter melts\textsubscript{I}}), while in languages such as Italian only the intransitive form is morphologically marked (\textit{scioglie\textsubscript{T}} vs. \textit{\textbf{si} scioglie\textsubscript{I}} `melts'). Conversely, in languages like Mongolian, the transitive form bears overt marking (\textit{xajl-\textbf{uul}-ax\textsubscript{T}} vs.  \textit{xajl-ax\textsubscript{I}} `melts'). In Japanese, both alternants are morphologically marked (\textit{atum-\textbf{eru}\textsubscript{T} } vs. \textit{atum-\textbf{aru}\textsubscript{I}} `gather’). 
These differences are reflected in the internal model's representations, as the morphological marking affects the tokens and, consequently, the models' internal representations.




Morphological paradigms can be considerably complex, featuring larger inventories of voices—such as passive forms—and, consequently, a greater number of morphological markers. In this respect, Turkish provides a clear example of a system with transparent inflectional morphology, realized as a set of allomorphs attached to the verbal root in the form of affixes \citep{oflazer-1993-two,Kornfilt1997, GokselKerslake2005,key2013morphosyntax}.

Modern Hebrew exhibits a similarly complex verbal paradigm, but one characterized by a different type of morphological organization. Its well-studied patterns, known as \textit{binyanim}, regulate how roots combine with morphological material to express a range of meanings, including causality \citep{McCarthy1979,Arad2005,Tsarfaty2004}. Although our data consist of Hebrew text without \textit{niqqud}—the diacritic signs indicating vowels—some of the voices still display non-concatenative morphology.\footnote{Not all roots permit all templatic structures \citep{kastner2019templatic}. While roots typically convey a single overarching semantic field, the relationships between forms in a given paradigm are not always transparent (e.g., for the root P\d{K}D, 'ordering'/'depositing'). This does not affect our investigation, which focuses on the morpho-syntactic aspects of the templates. Traditional grammars discuss seven binyanim, but in this paper we focus on four, following \citet{kastner2019templatic}.}

\begin{figure}
    \centering
    \includegraphics[width=1\linewidth]{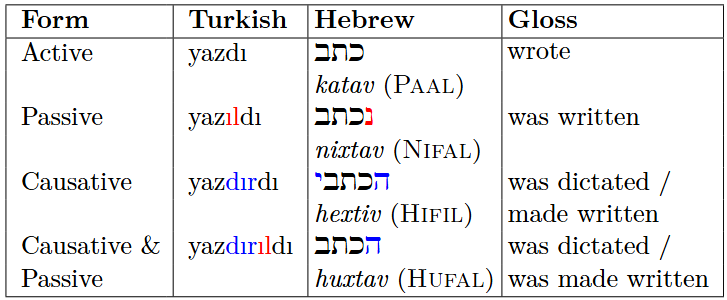}
    \caption{Verbal paradigm voices under investigation and relative examples for the Turkish verb \textit{yaz}- and the Hebrew root KTB (related to the act of writing). Hebrew binyanim are adapted from \citealt[574-575]{kastner2019templatic}, in brackets the name of the binyanim.}
    \label{fig:verbforms}
\end{figure}

The Turkish and Hebrew verbal paradigms under investigation are illustrated in Figure \ref{fig:verbforms}. In Turkish the active voice corresponds to the labile form of the inflected verb (e.g. person, number, tense features). The passive and causative voices are morphologically marked with affixes, while the causative-passive form combines both affixes.
In Hebrew, the Paal binyan represents the basic, labile transitive form (cf. Turkish active). Nifal corresponds to the passive voice, often—but not always—marked by a prefix containing the character nun \citep[71]{coffin2005reference}. Hifil and Hufal represent the causative active and passive voices, respectively; both can involve a prefix with the character \textit{he}, but the active form is disambiguated by the presence of a \textit{yod}—which marks a voiced palatal approximant and exemplifies non-concatenative morphology—alongside contextual cues. 

Morphosyntactic cues are readily available to speakers \citep{FRUCHTER201581}, but it remains unclear how they are captured by language models. Tokenization plays a key role in shaping the internal representations of language models \citep{hopton-etal-2025-functional}. 
However, standard substring tokenization could lead to different outcomes \citep{sennrich-etal-2016-neural,wu2016google}, depending on the  granularity of the process and its coherence with the morphological distinctions. In one alternative, 
the verbal form might be tokenized in a linguistically-congruent way, clearly separating roots and inflections. In another option, it might be split into small sub-morphemes --- sometimes as small as the grapheme level --- fragmenting all the overt linguistic elements to the point where their relationship is no longer fully represented. In yet another possibility, the verbal form can be handled as one  unit, keeping morphemes intact but blurring the distinction between root and inflection, and losing the compositional nature of morphological paradigms.

These alternative tokenization strategies determine what kind of morphological information is made available to the model directly and what morphological information needs instead to be induced in the hidden representations internal to the model. If a form is fragmented at the character level, the relation between root and pattern becomes opaque, as morphemes are broken into units too small to capture their function. If, instead, the morphemes appear in the same token, their internal structure is hidden. This motivates the use of a paradigm-level evaluation: by examining how models represent and process entire sets of related forms in complex settings, we can test whether morphological regularities survive tokenization choices and are encoded in sentence representations. The way words are tokenized affects their internal representations in sentences.

In this paper, we ask: Can current language models capture morphologically complex alternations in verbal paradigms in their internal representations, and how does tokenization affect their ability to represent these regularities?

To answer this question, we create structured datasets consisting of natural data (extracted from large-scale corpora) and synthetic data for a task appropriate for paradigms, the Blackbird Language Matrices (BLM) task \citep{merlo2023blackbirdlanguagematricesblm}, which is discussed in detail in Section \ref{blm-task}. We evaluate the representations generated by transformer models using this task. The BLM task is paradigm-based and has been shown to be challenging, aiming to capture core morphosyntactic and semantic abilities of language models \citep{nissim2025challengingabilitieslargelanguage}.

\section{The task}
\label{blm-task}
Blackbird Language Matrices (BLMs) are linguistic puzzles that implicitly describe paradigmatic linguistic systems \citep{merlo2023blackbirdlanguagematricesblm,merlo2022blackbirds,merlo2023-findings,an-etal-2023-blm,samo-etal-2023-blm,nastase2024exploring1,nastase2024exploring2,jiang-etal-2024-blm}. The task consists of a multiple-choice selecting the sentence that satisfies an underlying linguistic rule within a template. It has two components: (i) a context set of sentences that implicitly provides the information necessary to complete the linguistic paradigm, and (ii) an answer set of minimally differing contrastive sentences, where only one—the missing element in the pattern—is correct. 

\begin{figure*}
    \centering
\includegraphics[width=1\linewidth]{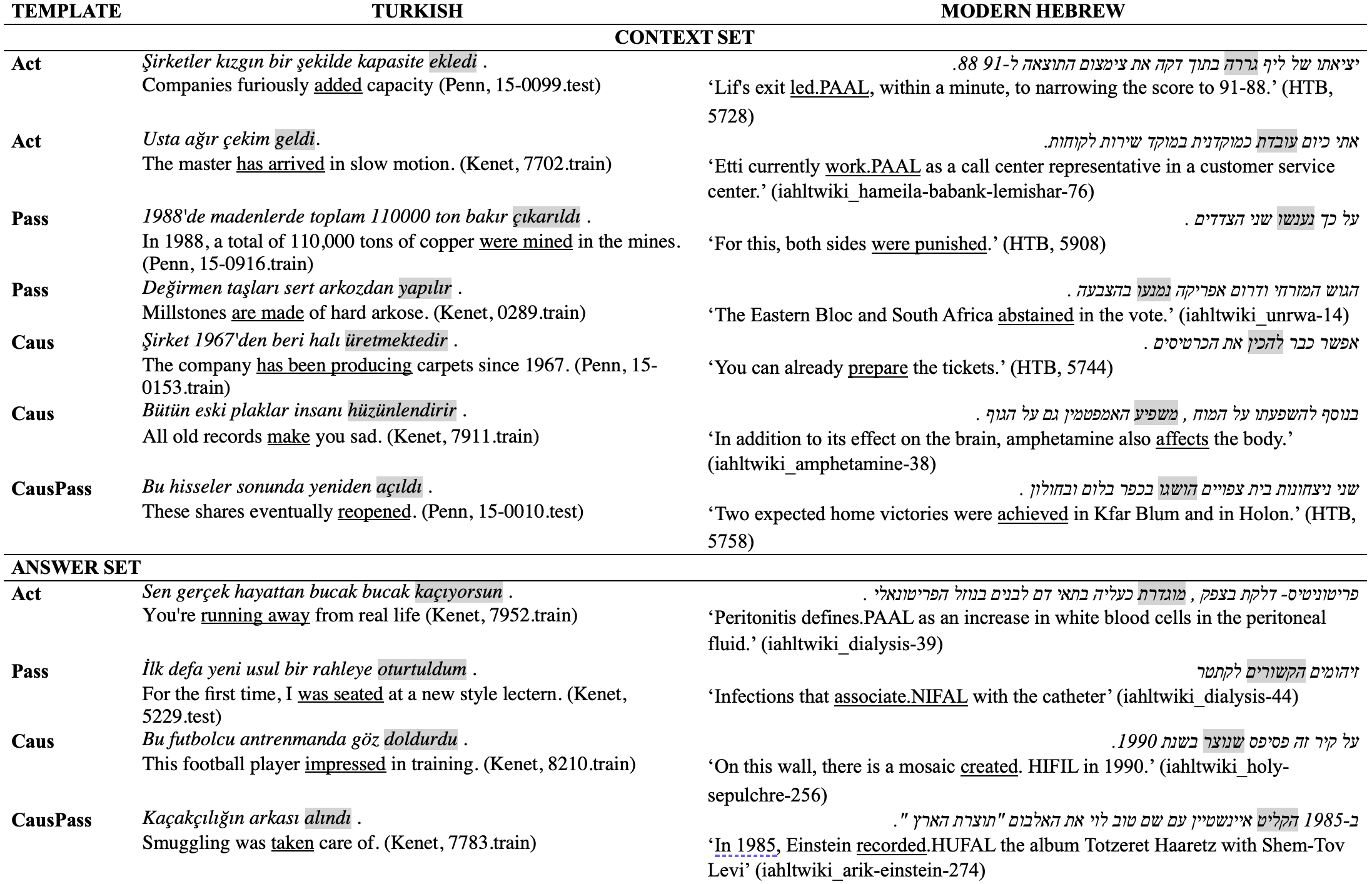}
\caption{BLM Template and instantiation in Turkish and Hebrew. The verb under investigation is underlined in the English translation. The indicated voice label is used only for error analysis, and not for training. The ID of the sentences refer to the dataset where the natural data are extracted as discussed in Section \ref{data}.}
\label{fig:blm-template}
\end{figure*}

By analyzing sentence continuations that follow specific syntactic or semantic patterns, BLMs serve as an investigative tool for identifying systematic morphological and syntactic regularities. They provide an informative setup for studying how internal representations encode knowledge of linguistic paradigms, making them suitable for complex cases, such as the Turkish regularly compositional strings, and the Hebrew binyanim system, where verbal alternations are not easily distinguishable. To learn these alternations, the model must observe all variants and capture the relations among them, representing the compositional \citep{oflazer-1993-two} or templatic structure of the system \citep{bobaljik2015suppletion}.
Templates are then instantiated creating curated datasets, and the task is then performed on sentence embeddings to test how linguistic information is encoded in the internal representations  of language models. An example of a BLM template and its instantiation in Turkish and in Hebrew is provided in Figure \ref{fig:blm-template}. Details on the template and data instantiation are given in Section \ref{data}.

Previous BLM studies on concatenative languages have examined agreement in Romance and English \citep{an-etal-2023-blm,nastase2024exploring1} and verb alternations such as the English \textit{spray/load} alternation \citep{samo-etal-2023-blm} or the English and Italian causatives and object-drop phenomena \citep{nastase2024exploring2}. These studies show that representations rely heavily on morphological cues, yielding excellent task performance. However, they also depend on superficial, character-level signals, which hinders correct transfer across languages \citep{nastase2024exploring2}. Character-level signals are integral to phenomena such as the prototypical endings in agreement inflection, the restricted use of prepositions in \textit{spray–load} constructions, and unique morphological markers like \textit{si} in the intransitive form of Italian causatives.

\section{Data and Models}

\label{data}
In this section, we describe our structured, curated datasets, and its construction.\footnote{Data is available at \url{https://www.idiap.ch/en/scientific-research/data/blm-caush} and \url{https://www.idiap.ch/en/scientific-research/data/blm-caust}.} We also introduce the language models under investigation and their tokenization strategies.

\subsection{BLM template}

Each BLM template is composed of a context sentence set and an answer set. The context  comprises three complete pairs of sentences with verbs inflected for a verbal form. The fourth pair is incomplete: one sentence illustrates the remaining form and the task consists in guessing the missing sentence. The answer set is composed of  four sentences, each one illustrating a different form. 
Figure \ref{fig:blm-template} shows an instantiation of the BLM template.\footnote{The presented examples are shown for illustrative purposes. At testing,  to avoid potential order effects, the order of presentation of the context set pairs (apart from the last one) rotates over all possible orders of context for each possible correct answer.}

\subsection{Instantiation}

Our dataset was created with natural occurring sentences extracted from treebanks annotated under the schema of the Universal Dependencies (UD; \citealt{nivre2015towards,de2021universal}). 
We retrieved the sentences using \textit{grew.match.fr}, by querying a simple variable X with the relevant annotation.
The data for Turkish are collected from news and non-fiction sources (Penn v. 2.16\footnote{\url{https://universaldependencies.org/treebanks/tr\_penn/index.html}}; 183,555 tokens, 16,396 trees) and grammar and dictionary examples (Kenet v. 2.16\footnote{ \url{https://universaldependencies.org/treebanks/tr\_kenet/index.html}}; 178,658 tokens, 18,687 trees). The query collects sentences where the main verb is annotated with the \textsc{Voice} parameter.\footnote{The queries are as follows,  \textsc{Act}: \textit{pattern}  X [upos="VERB"]; X [VerbForm = "Fin"] \textit{without}  X [Voice = "CauPass" | "Cau" | "Pass" | "Rcp" | "Rfl"]; \textsc{Pass}: \textit{pattern}  X [upos="VERB"]; X [Voice = "Pass"]; X [VerbForm = "Fin"]; \textsc{Caus}: \textit{pattern}  X [upos="VERB"]; X [Voice = "Cau"]; X [VerbForm = "Fin"]; \textsc{CausPass}: \textit{pattern}  X [upos="VERB"]; X [Voice = "CausPass"]; X [VerbForm = "Fin"].} The Modern Hebrew data were extracted from two treebanks of Hebrew containing respectively news (HBT v.2.15, \citealt{tsarfaty2013unified,mcdonald2013universal}; 114,648 tokens, 6,143 trees) and encyclopaedic entries (IAHLTWiki v. 2.15, henceforth IW; \citealt{ZeldesHowellOrdanBenMoshe2022}; 103,395 tokens; 5,039 trees). The query collects sentences where the main verb is annotated with relevant the morphosyntactic property \textsc{HebBinyan} for Hebrew.\footnote{The queries for Hebrew are as follows, \textsc{Act}: \textit{pattern} X [HebBinyan = "PAAL"]; \textsc{Pass}: \textit{pattern} X [HebBinyan = "NIFAL"]; \textsc{Caus}: \textit{pattern} X [HebBinyan = "HIFIL"]; \textsc{CausPass}: \textit{pattern} X [HebBinyan = "HUFAL"].}

These BLM-datasets contain the most complex type of lexical variation for the instantiation of the BLM template—sentences in the same BLM sequence do not share a common, limited lexicon within one BLM instance (indicated as ``type III'' or ``MaxLex'' in other work on BLMs \cite{an-etal-2023-blm,samo-etal-2023-blm,nastase2024exploring1,nastase2024exploring2}.

\subsection{Models}
To study the effect of tokenisation, we use sentence embeddings derived from both monolingual and multilingual models.
We use the monolingual sentence embeddings of BERTurk for Turkish (\textit{dbmdz/bert-base-turkish-cased}) and AlephBERT (\textit{onlplab/alephbert-base}) for Hebrew. 
Following previous work on verb alternations \citep{yi-etal-2022-probing, samo-etal-2023-blm, nastase2024exploring1}, we also use the sentence embeddings of the multilingual model Electra (\textit{google/electra-base-discriminator}, henceforth Multilingual/Multi).

We chose to work with transformer models rather than larger language or generative models for two reasons. First, transformers provide easy control over tokenization and embeddings. Second, using transformers allows us to obtain comparable, purely monolingual representations; by contrast, even large language models that can be described as monolingual do contain substantial amounts of English in their training data \citep{orlando-etal-2024-minerva}.

\section{Tokenisation} 
Tokenization plays a central role in shaping the internal representations of language models, making it essential to understand its impact on the modeling of linguistic phenomena. Most language models rely on sequential subword tokenization methods, such as Byte-Pair Encoding \citep{sennrich-etal-2016-neural} or WordPiece \citep{wu2016google}, which have been proven effective for concatenative morphologies \citep{hopton-etal-2025-functional}. 

As mentioned in the introduction, standard substring tokenization \citep{sennrich-etal-2016-neural,wu2016google} can yield very different outcomes depending on its granularity and alignment with linguistic structure. A verbal form may be segmented in a morphologically meaningful way, split into very small sub-morphemic units that obscure functional relationships, or treated as a single token that preserves the surface form while hiding internal compositionality. We present our results in light of the tokenization strategies for the verbal form, which represent the primary focus of our investigation.


\begin{figure}
    \centering
    \includegraphics[width=1\linewidth]{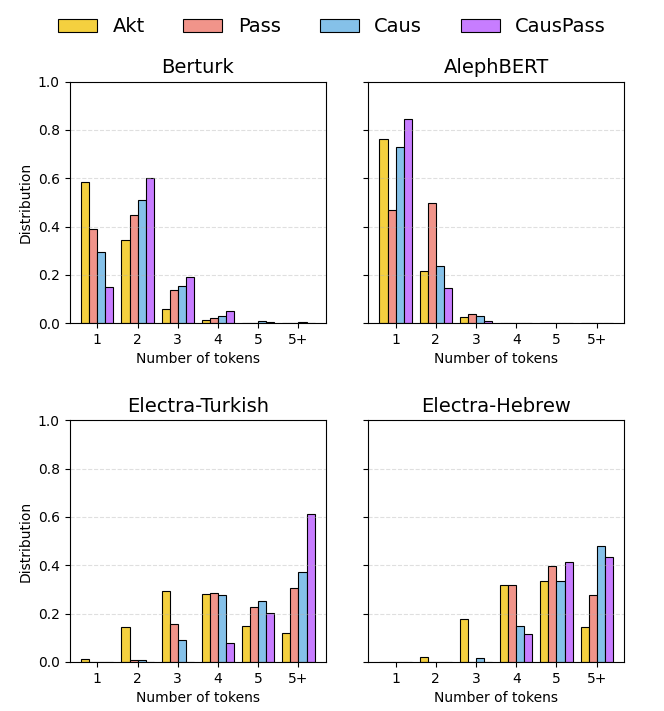}
    \caption{Number of tokens per voice forms across models and languages.}
    \label{fig:number-of-tokens}
\end{figure}

Figure \ref{fig:number-of-tokens} shows the number of tokens for each form. Monolingual models exhibit a similar pattern in both languages, but display a more atomic tokenization strategy for Hebrew verbal forms (an average of 1.329 tokens per form) than for Turkish (1.865 tokens per instance)\footnote{Table \ref{tab:tokenization} in the Appendix provides details on tokenization for all datasets under investigation.}. In Turkish, however, the causative-passive form is the most segmented, showing more tokens on average than the other marked forms such as causative or passive.

For Hebrew, the multilingual Electra model exhibits a more character-based representation (5.136 tokens per verbal form, 1.001 characters per token). This difference, particularly for Hebrew script data, can be traced to the type of training data: Electra’s fixed-size token vocabulary is dominated by frequent Latin-script languages, leaving only space for character-level representations of Hebrew \citep{muller-etal-2021-unseen, ahia-etal-2023-languages}. On the other hand, the multilingual model in Turkish does not fully adopt a character-based tokenization (1.950 characters per token). Nevertheless, the number of tokens per instance increases with morphological complexity: 3.825 tokens for the labile (active) form, 4.911 and 5.222 for passive and causative forms, respectively, and 6.868 when the latter two markings are combined.

\begin{table}[]
\centering
\footnotesize
\begin{tabular}{lll}
\textbf{Voice} & \textbf{Berturk} & \textbf{Electra} \\
\hline
Passive &
\#\#madı (91) &
\#\#ı (215) \\
&
\#\#du (80) &
\#\#r (68) \\
&
\#\#di (60) &
\#\#ld (54) \\
\hline
Causative &
\#\#dı (71) &
\#\#ı (1570) \\
&
\#\#di (49) &
\#\#r (455) \\
&
\#\#ıyor (44) &
\#\#yo (288) \\
\hline
Causative-Passive &
\#\#ildi (20) &
\#\#ı (1061) \\
&
\#\#ıldı (17) &
\#\#r (426) \\
&
\#\#ılıyor (8) &
\#\#yo (273) \\
\hline
\end{tabular}
    \caption{Turkish voices, top-3 most frequent tokens and their raw counts on the dataset (to be compared across models).}
    \label{tab:quality-of-tokens}
\end{table}

Finally, the linguistic informativeness of the tokens can be evaluated using the metrics in Table \ref{tab:quality-of-tokens} for the three marked forms (Passive, Causative and Causative-Passive). Table \ref{tab:quality-of-tokens} shows that the most frequent tokens in Turkish differ in quality between the monolingual and multilingual models for the marked forms. The table highlights not only differences in token quality but also in token length, with the multilingual model producing smaller subword units. In particular the most frequent tokens in the monolingual models are marker of tense and agreement (e.g. past tense \textit{-dı} or progressive \textit{-ıyor}). These results suggest that, in the more atomic monolingual model, the markers for passive and causative forms tend to remain attached to the root, while the multilingual model often splits them into separate subword tokens.

\section{Experiments}
\label{experiments}

We explore the behavior of a simple model through a series of experiments, reporting F1 scores and an error analysis.

\subsection{Materials \& Methods}

\paragraph{Data}

Each dataset contains 8000 instances, split into 90:10 training:testing. We use disjoint sets of training and testing instances. In both datasets, the answers are equally distributed for each voice (1800 training: 200 testing).  We ran four experiments for each dataset, isolating training and testing in answering each target voice.\footnote{A pilot study of three runs on each sub-dataset showed no differences in performance and errors across runs.} Each run used 50 training epochs.\footnote{All data for Hebrew were input as Hebrew alphabet characters without niqqud.}

\paragraph{System}

We used a Feed-Forward Neural Network (FFNN) as presented and discussed in the previous literature \citep{samo-etal-2023-blm}.  By using a feed-forward neural network (FFNN), we test whether the semantic relations targeted by the task can be captured from the input representation. We aim to keep the system simple so that no other complex variables could explain the results. This is particularly appropriate in the context of this type of BLM task, since the two languages differ in how broad relational patterns are expressed: in Turkish, these relations are more transparently compositional, whereas in Hebrew, they are less so. For each sentence in the BLM, we use the averaged token embeddings. The FFNN takes in the stacked embeddings, uses a max-margin loss in training and selects the answer that has the highest cosine similarity to the output.

By applying a structure-agnostic architecture that operates over the entire input simultaneously, the FFNN allows us to test whether the semantic relations targeted by the task are recoverable directly from the geometry of the embedding space. This is particularly appropriate in the context of this type of BLM, where the challenge lies in identifying broad relational patterns rather than sequential dependencies.

\subsection{Results}
\label{subsec:results}

\begin{figure}[h!]
    \centering
    
    \includegraphics[width=0.95\linewidth]{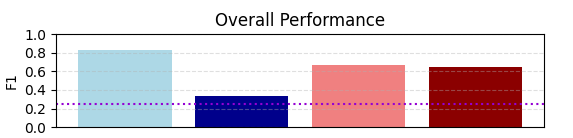}

    \includegraphics[width=0.46\linewidth]{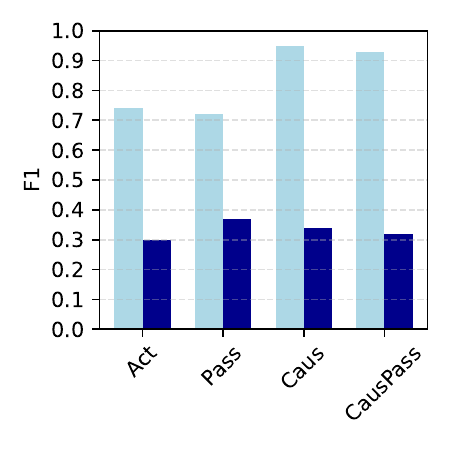}
    \includegraphics[width=0.46\linewidth]{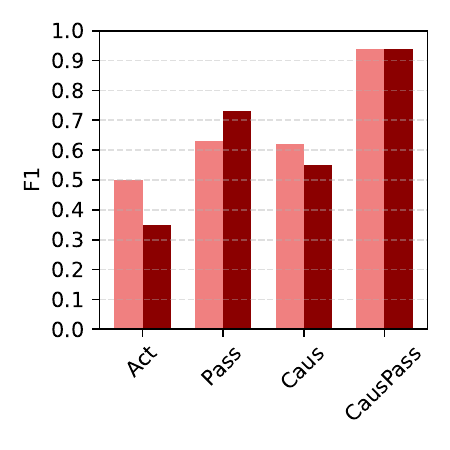}

    \centering
    \includegraphics[width=0.85\linewidth]{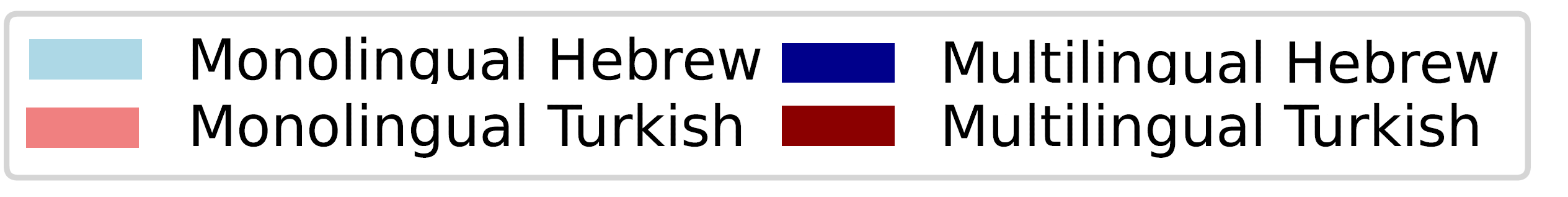}

    \caption{F1 for each voice as a correct answer across models. The dark violet dotted line in the upper panel indicates chance level.}
    \label{fig:figure_F1}
\end{figure}

Performance in terms of F1 scores is visualized in Figure \ref{fig:figure_F1}. 
The overall performance is similar across models for Turkish. In Hebrew, however, the monolingual model (average F1 score: 0.835) significantly outperforms the multilingual model (0.333), with a large difference (Mann–Whitney U test: $U = 16.0$, $p = .029$, $r = 1.0$).

For Turkish, both the monolingual and multilingual models behave similarly: the active form is the most difficult to predict, while marked forms are easier—particularly the causative passive, which contains double marking. However, the linguistic quality of the tokenization discussed in section \ref{data} does not introduce asymmetries. In Hebrew, the monolingual model performs better on the causative forms, but not consistently across all other marked forms such as passive. For the Hebrew multilingual model—which uses character-based tokenization—the binyanim in sentences are difficult to distinguish, resulting in low performance close to chance level.

\begin{figure}
    \centering
    \includegraphics[width=1\linewidth, trim=0 0 8cm 0,clip]{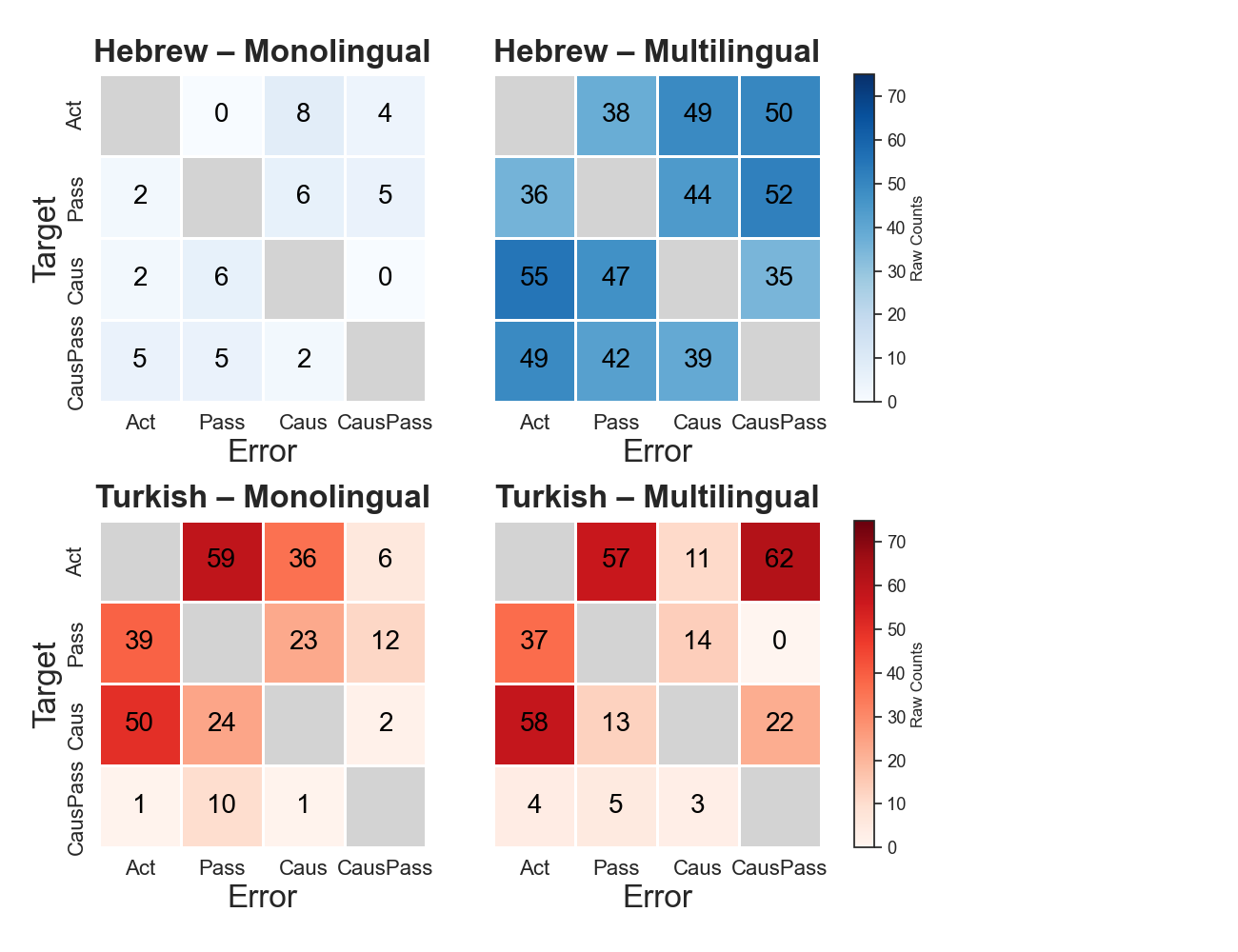}
    \caption{Confusion matrices of raw counts (test set \textit{n} = 200)}
    \label{fig:error-ea}
\end{figure}

Errors are visualized in the heatmap in Figure \ref{fig:error-ea}. We do not observe a consistently favoured answer across datasets and models except that in the Turkish monolingual, Passive is a prominent error when the target answer is Causative-Passive. For Hebrew the multilingual model shows a more distributed pattern of errors, consistent with chance-level performance. These results may indicate that granularity of tokenisation directly affects paradigm identification, at least for Hebrew. To better isolate the role of verbal paradigms, we create a synthetic dataset with reduced sentence complexity.

\subsection{Analyzing the verbal paradigm}
We created a second -- more synthetic -- dataset, which we label \textsc{VerbOnly}. This dataset contains only the verb corresponding to each sentence in the previous dataset, with no additional lexical content. The resulting sentences are grammatically correct, as both Hebrew and Turkish allow both subject and object drop \citep{vainikka1999empty,erteschik2013missing,meral2014silent}.\footnote{The automatically retrieved inflected isolated forms in Hebrew may also contain affixes representing prepositions, determiners or complementizers (see also \citealt{shmidman-rubinstein-2024-computational}).}

Specifically, this setup abstracts away the voices from both syntactic and lexical context, allowing us to focus exclusively on the verbal form and its morphological information. While this ultimately simplifies the task — since each sentence now contains only the verb, reducing its length and possibly noise — it allows for a more direct analysis of paradigms in the strict sense.

\begin{figure}[h!]
    \centering
    
    \includegraphics[width=0.95\linewidth]{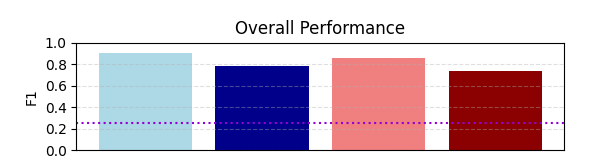}

    \includegraphics[width=0.46\linewidth]{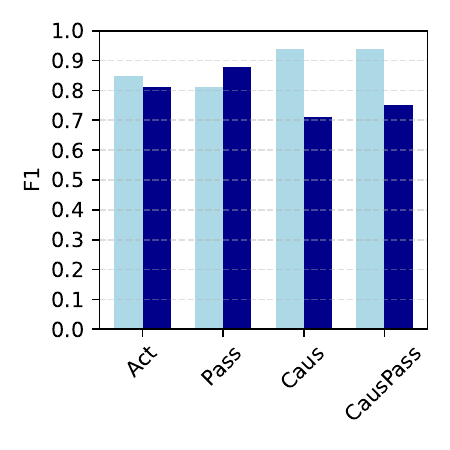}
    \includegraphics[width=0.46\linewidth]{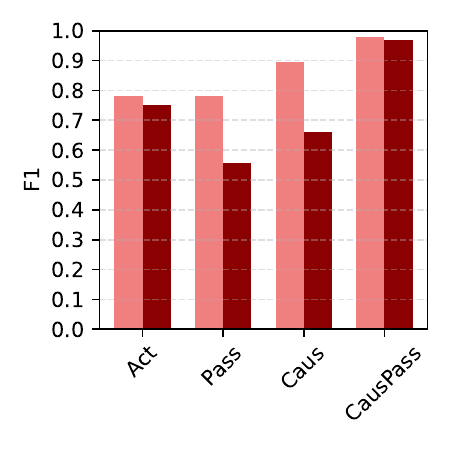}

    \centering
    \includegraphics[width=0.85\linewidth]{figures/legends.png}

    \caption{F1 for each binyan as a correct answer across models for the \textsc{VerbOnly} dataset. The dark violet dotted line indicates chance level.}
    \label{fig:F1-verbonly}
\end{figure}

We run the same experiment on this synthetic dataset. Results are shown in Figure \ref{fig:F1-verbonly}.

In Turkish we observe an overall improvement of the performance, with the monolingual model performing slightly better than the multilingual in every voice. In Hebrew, the monolingual model is consistent with excellent results. The multilingual model also shows improved performance, approaching the monolingual model, although it still lags behind with respect to the causative forms.

\begin{figure}
    \centering
    \includegraphics[width=1\linewidth, trim=0 0 8cm 0,clip]{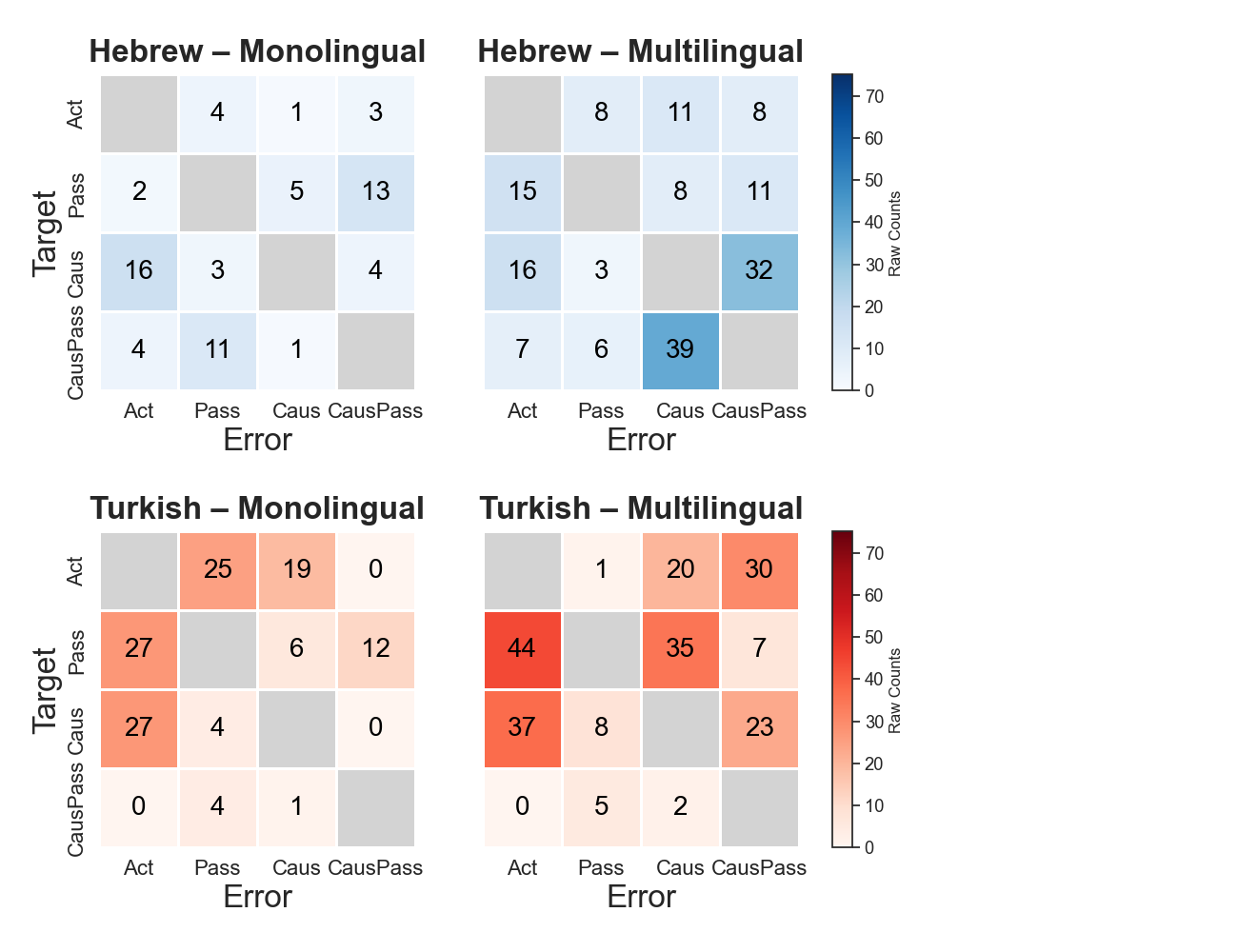}
    \caption{Confusion matrices of raw counts (test set \textit{n} = 200) for the \textsc{VerbOnly} dataset.} 
    \label{fig:error-analysis-verbonly}
\end{figure}

As Figure \ref{fig:error-analysis-verbonly} shows, in Turkish we observe a similar distribution of errors as the dataset containing full sentences. In Hebrew, however, one particularly informative type of error involves the confusion between Caus and CausPass, which share common morphological elements. Notably, in the multilingual \textsc{VerbOnly} model, which uses character-level tokenization, the most frequent error for target CausPass is Caus (\textit{z} = 4.31, \textit{p} $<$ .01), and vice versa 
(\textit{z} = 5.35, \textit{p} $<$ .01). This confusion suggests that the model’s character-based representations capture surface-level morphological similarity and fail to fully distinguish deeper differences between the two binyanim.

\subsection{Discussion}
\label{discussion}
Our results provide a clear answer to our research question: transformer models can indeed capture morphologically complex alternations in their internal representations, but this capacity is highly dependent on how tokenization interacts with the specific morphological structure of a language. For Turkish—a language with transparent, concatenative morphology—both monolingual and multilingual models succeeded. The monolingual model employed a more atomic tokenization, often representing entire inflected word forms as single tokens, while the multilingual model used a more fragmented subword segmentation. Crucially, both strategies proved effective, indicating that for agglutinative systems, explicit surface forms—whether present in the atomic representation or in smaller segments—provide sufficient cues to infer paradigmatic relationships.

Conversely, the exact form of  tokenization for Hebrew's morphology becomes decisive. The multilingual model based on character-level segmentation  failed to capture the templatic binyanim system in natural sentences, performing near chance. Its overly fragmented tokenization may hinder the systematic co-dependence between root and pattern, rendering the paradigm opaque. The performance of the multilingual improves on simplified data, but it still lags behind the monolingual model.

These findings show that tokenization is not merely a preprocessing step, but possibly a linguistic filter that determines which morphological regularities are learnable. For the Turkish systems, both atomic and segmented tokenizations can be effective, but for Hebrew morphology, a representation that preserves or contains the integrity of the morphological form—whether atomic or appropriately segmented—is critical. The BLM task proves effective in diagnosing this interplay between tokenization strategy and morphological typology.

\section{Related Work}
\label{related}
The creation of paradigm-based datasets is useful for evaluating the generalization capacity of language models, for instance, across different morphological patterns \citep{batsuren-etal-2022-unimorph, sigmorphon-2024, warstadt-etal-2020-blimp-benchmark}. Verb alternations, especially in English, have been the object of recent investigation in LLMs showing excellent performance \citep{kann-etal-2019-verb,warstadt-etal-2019-investigating,wilson2023abstract}. \citet{yi-etal-2022-probing} suggest that LLMs with contextual embeddings capture linguistic information about verb alternation classes at both the word and sentence levels in English. Also the semantic properties of the argument of verbs (agents, patients) have been tested with transformer models  \citep{proietti-etal-2022-bert}. 

The evaluation of LLMs’ linguistic competence often relies on benchmark suites that make use of synthetic datasets. Synthetic datasets are constructed to probe specific grammatical phenomena in a controlled manner, frequently using minimal pairs or carefully designed paradigms \citep{warstadt-etal-2019-investigating, warstadt-etal-2020-blimp-benchmark}. While automatic generation facilitates large-scale evaluation, it also raises concerns regarding distributional biases \citep{zhang2025doestrainingsyntheticdata, nadas-etal-2025-survey, griffiths2024bayes}. In this paper, we begin our analysis by extracting data from large-scale naturalistic datasets, which may provide a more faithful basis for evaluating language models’ representations of natural language \citep{jumelet-etal-2025-multiblimp}.

The way tokenization is implemented \citep{rajaraman2024theorytokenizationllms} also influences the accuracy of classification tasks in language identification and/or (neural) machine translation \citep{kanjirangat-etal-2023-optimizing,domingo2019doestokenizationaffectneural}. Approaches using neural encoders that operate directly on character sequences have been proposed and discussed \citep{clark-etal-2022-canine}. \citet{hopton-etal-2025-functional} demonstrated that subword tokenization can distinguish between function words (e.g., those indicating verbal constraints) and content words even in low-resource languages without annotated data. 

Turkish has long been regarded as a key language for computational linguistics research on the interaction between morphology and tokenization, due to its highly agglutinative structure and productive inflectional system \citep{ataman2017linguisticallymotivatedvocabularyreduction,ataman-federico-2018-evaluation}. \citet{Toraman2023TokenizationTurkish} show that morphology-level tokenization for Turkish performs competitively with standard subword methods. Similarly, morphology-aware tokenization has been discussed in processing Semitic languages, proposing new tokenization algorithms \citep{goldman2022morphology}, such as linguistically informed extensions of BPE \citep{asgari2025morphbpemorphoawaretokenizerbridging}.  \citet{gueta2023explicitmorphologicalknowledgeimproves}  integrate morphological knowledge directly into pretraining via specialized tokenization, showing gains on Hebrew across semantic and morphological benchmarks. According to \citet{dang2024tokenizationmorphologymultilinguallanguage}, both languages capture morphological knowledge effectively across various tokenization strategies, such as (sub)word-level and character-level approaches.




\section{Conclusions}
\label{conclusions}
In this study, we examined how tokenization affects the ability of language models to represent complex verbal paradigms in Turkish and Hebrew. Our experiments show that tokenization granularity does interact with how internal sentence representations capture morphologically-complex alternations. Our results show that overall, monolingual models perform better than multilingual ones, indicating that performance is high when morphemes remain intact, whereas fragmentation can obscure systematic relations. Overall, these results underscore the importance of paradigm-level evaluation for understanding how models encode linguistic knowledge and highlight that tokenization strategy and language-specific morphology jointly shape internal representations. Future work should explore linguistically-informed tokenization schemes and extend these analyses to other morphologically-rich languages as well as to other linguistic phenomena to better understand the interaction between tokenization and linguistic knowledge in models.
\section*{Limitations}

Future work could address the limitations of this contribution by expanding language coverage, exploring additional models and architectures, and performing comprehensive validation, as well a human upperbound.

\section*{Ethics}
We used datasets derived from publicly available corpora, which may include content such as news articles and other publicly accessible materials. It is important to note that these datasets may contain sensitive or potentially upsetting topics. We acknowledge that such content may be distressing to some individuals. We encourage users to approach the results with awareness of these considerations.

\section*{Acknowledgments}
We gratefully acknowledge the partial support of this work by the Swiss National Science Foundation, through grant SNF Advanced grant TMAG-1\_209426 to PM. We thank Gökhan Özbulak and Ur Shlonsky for precious comments on the data.

\bibliography{custom}

\newpage

\appendix
\section*{Appendix}

\begin{table*}[]
\scriptsize
\centering
\begin{tabular}{r|r|rrrr|rrrrrr}
  &  &    \multicolumn{4}{c|}{\textbf{\textsc{Sentences}}}               & \multicolumn{5}{c}{\textbf{\textsc{Verbs}}}                 \\ \hline
\textbf{\textsc{Voice}}     &    \textbf{\textsc{Inst}}  & \textbf{\textsc{Ch}}   & \textbf{\textsc{Tok} }   & \textbf{\textsc{tok/inst}} & \textbf{\textsc{ch/tok}} & \textbf{\textsc{Ch}}    & \textbf{\textsc{Tok}}  & \textbf{\textsc{tok/inst}}  & \textbf{\textsc{ch/tok}} & \textbf{\textsc{1v-tok}}  \\ 
\hline 
\multicolumn{11}{c}{\textsc{Hebrew - Monolingual}}                                                             \\  \hline 
\textbf{Total}    & \textbf{6899} & \textbf{880210} & \textbf{215195} & \textbf{31.130}   & \textbf{4.093}  & \textbf{35140} & \textbf{8966}  &  \textbf{1.329}     & \textbf{3.888}   & \textbf{3746}  \\
Act      & 1928 & 242839 & 60117  & 31.181   & 4.039  &  8652  & 2254  & 1.169     & 3.839   & 1155  \\
Pass     & 1892 & 234444 & 57435  & 30.357   & 4.082  &  9516  & 2478  & 1.310     & 3.840   & 897   \\
Caus      & 1972 & 262825 & 63653  & 32.278   & 4.129  &  10815 & 2495  & 1.265     & 4.335   & 1104  \\
CausPass & 1107 & 140102 & 33990  & 30.705   & 4.122  &  6157  & 1739  & 1.571     & 3.541   & 590   \\ \hline 
\multicolumn{11}{c}{\textsc{Hebrew - Multilingual}}                                                      \\ \hline 
\textbf{Total}     & \textbf{6899} & \textbf{880210} & \textbf{711294} & \textbf{102.993} & \textbf{1.237}  & \textbf{35140} & \textbf{35101} & \textbf{5.136}    & \textbf{1.001}  & \textbf{169}   \\
Act      & 1928 & 242839 & 195663 & 101.485  & 1.241    & 8652  & 8630  & 4.476    & 1.003  & 49    \\
Pass     & 1892 & 234444 & 189327 & 100.067  & 1.238    & 9516  & 9512  & 5.027    & 1.000  & 41    \\
Caus      & 1972 & 262825 & 212857 & 107.940  & 1.235    & 10815 & 10802 & 5.478    & 1.001  & 40    \\
CausPass & 1107 & 140102 & 113447 & 102.481  & 1.235     & 6157  & 6157  & 5.562    & 1.000  & 39    \\ \hline         
\multicolumn{11}{c}{\textsc{Turkish - Monolingual}}                                                                  \\ \hline
\textbf{Total}    & \textbf{5112} & \textbf{387195} & \textbf{81522}  & \textbf{15.984}   & \textbf{4.784} &  \textbf{46379} & \textbf{8952}  &  \textbf{1.865}   & \textbf{5.391} & \textbf{3044}  \\
Act      & 1841 & 139698 & 29663  & 16.112   & 4.710  &  14002 & 2803  & \cellcolor{gray!15} 1.523    & 4.995  & 1341  \\
Pass     & 1879 & 138224 & 29006  & 15.437   & 4.765  &  17688 & 3377  & \cellcolor{gray!18} 1.797    & 5.238  & 870   \\
Caus     & 1240 & 97437  & 20435  & 16.480   & 4.768  & 12661 & 2442  & \cellcolor{gray!19}1.969    & 5.185  & 657   \\
CausPass & 152  & 11836  & 2418   & 15.908   & 4.895  & 2028  & 330   & \cellcolor{gray!21}2.171    & 6.145  & 176   \\ \hline 
\multicolumn{11}{c}{\textsc{Turkish - Multilingual}}                                                                \\ \hline 
\textbf{Total}    & \textbf{5112} & \textbf{387195} & \textbf{162347} & \textbf{32.148}   & \textbf{2.379}  &  \textbf{46379} & \textbf{23793} &  \textbf{5.207}   &  \textbf{1.950}  & \textbf{1847}  \\
Act      & 1841 & 139698 & 58299  & 31.667   & 2.396  &  14002 & 7042  & \cellcolor{gray!38}3.825    & 1.988  & 698   \\
Pass     & 1879 & 138224 & 57846  & 30.786   & 2.390  & 17688 & 9228  & \cellcolor{gray!49}4.911    & 1.917  & 480   \\
Caus      & 1240 & 97437  & 41199  & 33.225   & 2.365  &  12661 & 6479  & \cellcolor{gray!52}5.225    & 1.954  & 465   \\
CausPass & 152  & 11836  & 5003   & 32.914   & 2.366   & 2028  & 1044  & \cellcolor{gray!68}6.868    & 1.943  & 204  
\end{tabular}
    \caption{Comparison of AlephBERT, Berturk and ELECTRA tokenization of sentences and inflected verbs in terms of number of tokens (\textsc{Tok}), number of characters (\textsc{Ch}), token per word (\textsc{Tok/W}), characters per token (\textsc{Ch/Tok}) and one-token verbal forms (\textsc{1v-Tok}). The highlighted cells in grey for the tokenization of verbs refer to the increasing number of token pro verbal form.}
    \label{tab:tokenization}
\end{table*}
\end{document}